\documentclass[conference]{IEEEtran}
\usepackage{amsmath}
\usepackage{amssymb}
\usepackage{graphicx}
\usepackage{booktabs}
\usepackage{url}
\usepackage{hyperref}
\usepackage{CJKutf8}
\usepackage{tikz}
\usetikzlibrary{arrows.meta,positioning}
\usepackage{xcolor}

\title{DuplexChat: Constructing Speaker-Separated Full-Duplex Dialogue Speech at Scale for\\Spoken Dialogue Language Modeling}
\author{
\IEEEauthorblockN{Wataru Nakata$^{1,2}$, Yuki Saito$^{1,2}$ and Hiroshi Saruwatari$^{1}$}
\IEEEauthorblockA{
$^1$The University of Tokyo, Japan,
$^2$National Institute of Advanced Industrial Science and Technology, Japan.
\\
\{nakata-wataru855, sythonuk\}@g.ecc.u-tokyo.ac.jp
}
}

\begin{document}
\maketitle

\begin{abstract}
Full-duplex spoken dialogue models are trained on conversational speech in which each speaker is represented as a separate stream, but existing large-scale public speech corpora are mostly monaural, making them unsuited for SDLM training. We present \textit{DuplexChat}, an open-source corpus for full-duplex spoken dialogue models, and \textit{DuplexChat-Pipe}, a pipeline for constructing speaker-separated full-duplex dialogue speech from public podcast feeds. DuplexChat-Pipe filters language-specific podcast feeds, retrieves and cleans episode audio, extracts diarization-guided two-speaker dialogue clips, and applies speech separation and restoration to produce one channel per speaker. Running this pipeline yields a speaker-separated spoken dialogue corpus covering 282{,}634 hours of English and 132{,}723 hours of Japanese. Analysis results on DuplexChat show that it contains turn-taking dynamics present in human dialogues.
\end{abstract}

\section{Introduction}
Speech is a primary medium of human communication and a convenient interface for conversational AI. However, natural spoken interaction requires more than converting text responses into speech. Human conversation is tightly coordinated in real time. Speakers take turns with short gaps, provide backchannels, and frequently overlap~\cite{doi:10.1073/pnas.0903616106}. These conversational dynamics carry essential information that enables rich, responsive interaction, yet they are difficult to capture with conventional cascaded spoken dialogue systems~\cite{wang2025freezeomni}, which serialize automatic speech recognition, text generation, and speech synthesis~\cite{zhang-etal-2023-speechgpt}. Spoken Dialogue Language Models (SDLMs) such as dGSLM~\cite{nguyen-etal-2023-generative} and Moshi~\cite{defossez2024moshispeechtextfoundationmodel} address this limitation by modeling the dialogue in an end-to-end manner, making them a promising direction for more natural voice interaction.

As with text language models~\cite{kaplan2020scalinglawsneurallanguage}, SDLMs are expected to benefit from larger training corpora, yet the data they require is difficult to obtain at scale. Training requires full-duplex dialogue speech~\cite{nguyen-etal-2023-generative,defossez2024moshispeechtextfoundationmodel}: conversations in which the two participants are recorded as separate audio tracks, so that turn-taking, backchannels, and overlapping speech remain observable. Existing full-duplex resources are dominated by telephone-conversation corpora such as CallHome~\cite{ldc2008_cabank_callhome_english,ldc2008_cabank_callhome_german,ldc2008_cabank_callhome_chinese,ldc2008_cabank_callhome_spanish,ldc2008_cabank_callhome_japanese} and Fisher~\cite{cieri-etal-2004-fisher}, whose collection requires recruiting paired participants and therefore does not scale easily. In contrast, large public speech corpora such as GigaSpeech~\cite{chen21o_interspeech}, YODAS~\cite{10389689}, and the Spotify Podcast Dataset~\cite{clifton-etal-2020-100000} provide web-scale audio but consist of monaural recordings. Once two speakers have been mixed into a single channel, the speaker-specific timing needed for full-duplex modeling is no longer directly available. What is missing is an open and scalable method that combines web-scale audio with speaker-separated channels.

Scaling such a corpus beyond recorded corpora calls for harvesting audio from the Internet rather than collecting telephone conversations. Podcasts are a natural source to crawl: distributed through public RSS feeds, they supply an effectively unlimited and continually growing stream of spontaneous dialogue.
However, podcast audio is mixed-channel audio and interleaved with substantial non-dialogue material such as monologues, advertisements, and music. 

In this paper, we propose \textit{DuplexChat-Pipe}, an open-source pipeline that constructs full-duplex dialogue speech by crawling public podcast feeds, and the resulting corpus named \textit{DuplexChat}.
DuplexChat is a spoken dialogue collection comprising 282{,}634 hours of English and 132{,}723 hours of Japanese two-speaker dialogue. It is made available in a full-duplex speaker-separated format by using the  speech restoration and separation model. To our knowledge, DuplexChat-Pipe is the first open-source, re-runnable pipeline that produces \emph{speaker-separated} full-duplex dialogue speech at web scale, and the resulting DuplexChat ($\sim$415k\,h) is the largest dialogue resource to date (Table~\ref{tab:corpus-comparison}). Our contributions are as follows:
\begin{itemize}
  \item We present an open-source pipeline named DuplexChat-Pipe which produces speaker-separated spoken dialogue tracks for SDLM training.
  \item We release DuplexChat, a spoken dialogue corpus comprising of 282{,}634 hours of English and 132{,}723 hours of Japanese---by far the largest open resource for SDLM training.
\end{itemize}
The corpus is publicly available\footnote{\url{https://github.com/sarulab-speech/DuplexChat}}.

\begin{table*}[!t]
\centering
\caption{Comparison of representative public speech resources and DuplexChat. Sorted by corpus hours.}
\label{tab:corpus-comparison}
\footnotesize
\begin{tabular}{llclccc}
\toprule
Resource & Lang. & Source & Size [h]& Dialogue & Sep. channels & Pipeline release \\
\midrule
CallHome~\cite{ldc2008_cabank_callhome_english,ldc2008_cabank_callhome_german,ldc2008_cabank_callhome_chinese,ldc2008_cabank_callhome_spanish,ldc2008_cabank_callhome_japanese} & Multi & Telephone & 0.3k & Yes & Recorded & No \\
Fisher~\cite{cieri-etal-2004-fisher} & En & Telephone & 2k & Yes & Recorded & No \\
GigaSpeech~\cite{chen21o_interspeech} & En & Internet & 10k & Mixed & No & No \\
Spotify Podcast Dataset~\cite{clifton-etal-2020-100000} & En & Podcasts & 60k & Mixed & No & No \\
J-CHAT~\cite{jchat} & Ja & Podcasts+ YouTube & 76k & Yes & No & No \\
Emilia~\cite{emilia} & Multi & Internet & 101k & No & No & Yes \\
YODAS~\cite{10389689} & Multi & YouTube & 370k & Mixed & No & No \\
DuplexChat (ours) & En/Ja & Podcasts & 415k & Yes & Separated & Yes \\
\bottomrule
\end{tabular}
\vspace{-4mm}
\end{table*}

\section{DuplexChat-Pipe}
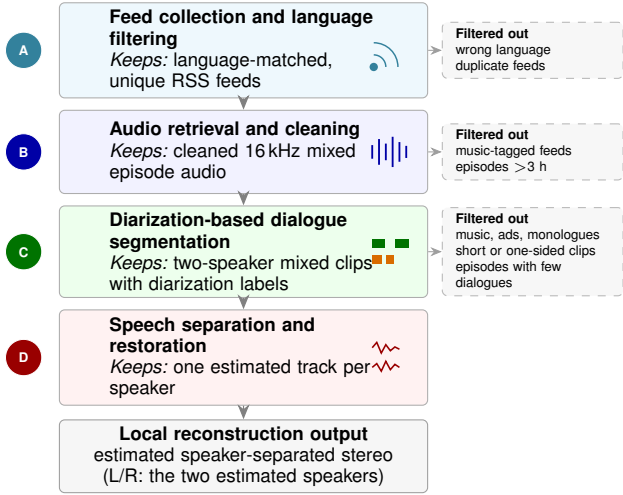
\begin{figure}[!t]
\centering
\begin{tikzpicture}[
  font=\scriptsize\sffamily,
  node distance=1.5mm,
  stage/.style={
    draw,
    draw=black!32,
    rounded corners=2pt,
    align=left,
    inner xsep=3.2pt,
    inner ysep=2.6pt,
    minimum width=0.55\columnwidth,
    minimum height=11mm,
    text width=0.40\columnwidth
  },
  reject/.style={
    draw=black!28,
    dashed,
    rounded corners=2pt,
    fill=gray!6,
    align=left,
    inner xsep=2.2pt,
    inner ysep=2.2pt,
    minimum width=0.27\columnwidth,
    text width=0.23\columnwidth,
    font=\tiny\sffamily
  },
  pill/.style={
    circle,
    draw=white,
    line width=0.35pt,
    minimum size=4.8mm,
    inner sep=0pt,
    text=white,
    font=\tiny\bfseries\sffamily
  },
  output/.style={
    draw=black!38,
    rounded corners=2pt,
    fill=gray!7,
    align=center,
    inner xsep=3pt,
    inner ysep=2.4pt,
    minimum width=0.55\columnwidth,
    text width=0.49\columnwidth
  },
  arrow/.style={-{Stealth[length=2mm,width=1.6mm]}, draw=black!50, line width=0.45pt},
  drop/.style={-{Stealth[length=1.5mm,width=1.2mm]}, draw=black!38, dashed, line width=0.35pt}
]
\node[stage, fill=cyan!6] (feeds)
  {\textbf{Feed collection and language filtering}\\\textit{Keeps:} language-matched, unique RSS feeds};
\node[pill, fill=cyan!55!black, left=2.2mm of feeds] {A};
\node[reject, right=1.8mm of feeds] (feedsout)
  {\textbf{Filtered out}\\wrong language\\duplicate feeds};

\node[stage, fill=blue!5, below=of feeds] (audio)
  {\textbf{Audio retrieval and cleaning}\\\textit{Keeps:} cleaned 16\,kHz mixed episode audio};
\node[pill, fill=blue!65!black, left=2.2mm of audio] {B};
\node[reject, right=1.8mm of audio] (audioout)
  {\textbf{Filtered out}\\music-tagged feeds\\episodes $>$3 h};

\node[stage, fill=green!7, below=of audio] (segment)
  {\textbf{Diarization-based dialogue segmentation}\\\textit{Keeps:} two-speaker mixed clips with diarization labels};
\node[pill, fill=green!45!black, left=2.2mm of segment] {C};
\node[reject, right=1.8mm of segment] (segmentout)
  {\textbf{Filtered out}\\music, ads, monologues\\short or one-sided clips\\episodes with few dialogues};

\node[stage, fill=red!5, below=of segment] (separate)
  {\textbf{Speech separation and restoration}\\\textit{Keeps:} one estimated track per speaker};
\node[pill, fill=red!60!black, left=2.2mm of separate] {D};

\node[output, below=1.8mm of separate] (stereo)
  {\textbf{Local reconstruction output}\\estimated speaker-separated stereo (L/R: the two estimated speakers)};

\coordinate (i1) at ([xshift=-5mm]feeds.east);
\fill[cyan!55!black] ([xshift=-2.2mm,yshift=-2.2mm]i1) circle (0.55mm);
\draw[cyan!55!black,line width=0.55pt] ([xshift=-2.2mm,yshift=-0.6mm]i1) arc[start angle=90,end angle=0,radius=1.6mm];
\draw[cyan!55!black,line width=0.55pt] ([xshift=-2.2mm,yshift=1.0mm]i1) arc[start angle=90,end angle=0,radius=3.2mm];
\coordinate (i2) at ([xshift=-5mm]audio.east);
\foreach \x/\h in {-2.4/0.9,-1.5/1.7,-0.6/1.1,0.3/2.0,1.2/1.3,2.1/0.8}{%
  \draw[blue!65!black,line width=0.7pt] ([xshift=\x mm,yshift=-\h mm]i2)--([xshift=\x mm,yshift=\h mm]i2);}
\coordinate (i3) at ([xshift=-5mm]segment.east);
\fill[green!45!black] ([xshift=-2.4mm,yshift=0.5mm]i3) rectangle ([xshift=-0.8mm,yshift=1.6mm]i3);
\fill[green!45!black] ([xshift=0.6mm,yshift=0.5mm]i3) rectangle ([xshift=2.4mm,yshift=1.6mm]i3);
\fill[orange!85!black] ([xshift=-2.4mm,yshift=-1.6mm]i3) rectangle ([xshift=-1.2mm,yshift=-0.5mm]i3);
\fill[orange!85!black] ([xshift=-0.6mm,yshift=-1.6mm]i3) rectangle ([xshift=0.4mm,yshift=-0.5mm]i3);
\coordinate (i4) at ([xshift=-5mm]separate.east);
\draw[red!60!black,line width=0.6pt] ([xshift=-2.4mm,yshift=1.3mm]i4) -- ++(0.6mm,0.7mm) -- ++(0.6mm,-1.4mm) -- ++(0.6mm,1.1mm) -- ++(0.6mm,-0.8mm) -- ++(0.6mm,0.5mm) -- ++(0.6mm,-0.3mm);
\draw[red!60!black,line width=0.6pt] ([xshift=-2.4mm,yshift=-1.3mm]i4) -- ++(0.6mm,0.5mm) -- ++(0.6mm,-1.0mm) -- ++(0.6mm,1.3mm) -- ++(0.6mm,-1.1mm) -- ++(0.6mm,0.7mm) -- ++(0.6mm,-0.4mm);
\draw[arrow] (feeds) -- (audio);
\draw[arrow] (audio) -- (segment);
\draw[arrow] (segment) -- (separate);
\draw[arrow] (separate) -- (stereo);
\draw[drop] (feeds) -- (feedsout);
\draw[drop] (audio) -- (audioout);
\draw[drop] (segment) -- (segmentout);
\end{tikzpicture}
\caption{Overview of  proposed DuplexChat-Pipe.}
\label{fig:pipeline}
\vspace{-4mm}
\end{figure}

Fig.~\ref{fig:pipeline} shows the proposed DuplexChat-Pipe. It proceeds in four stages: feed collection, audio retrieval and cleaning, diarization-based dialogue segmentation, and speech separation/restoration. We describe each in turn.

\subsection{Feed Collection and Language Filtering}
We start the crawl from PodcastIndex\footnote{\url{https://podcastindex.org/}}, a public index of podcasts that is already used as a crawling source in previous work~\cite{jchat}.
PodcastIndex makes its database publicly available, including the RSS feed URLs. We use These RSS feed URLs to retrieve podcast episode metadata and audio links. For each target language, we retain feeds whose language tag matches the requested locale and remove duplicate feed URLs before crawling. The output of this stage is a language-specific set of unique RSS feeds that contains links to candidate episode audio files.

\subsection{Audio Retrieval and Cleaning}
After RSS feed collection, we first apply feed-level metadata filtering to remove music shows. A feed is discarded when its category tags match a curated list of music-related keywords (e.g., song, music, rock, and pop). We then download the linked audio for each remaining episode and resample it to 16\,kHz. A second filter removes episodes longer than three hours, which in our crawl are predominantly music programs and long-form streams rather than two-speaker conversation. The output of this stage is a set of cleaned episode-level recordings; at this point, each episode is still a mixed-channel audio file containing all speakers and non-dialogue material that survived the metadata filters.

\subsection{Dialogue Segmentation}
Raw podcast audio contains substantial non-dialogue material, including music, advertisements, and monologues. We therefore segment each cleaned episode into two-speaker dialogue clips using labels from an open-source speaker diarization model\footnote{We used \texttt{pyannote/speaker-diarization-community-1}.}. Following previous work~\cite{jchat}, we split the audio at silence gaps of five seconds or longer and retain only spans in which exactly two speakers are active over the span, rejecting any span in which the other speakers are diarized. 
We then discard segments shorter than ten seconds and segments in which a single speaker accounts for more than 80\% of the diarization result, considering these samples as monologues. Segments longer than ten minutes are divided into segments shorter than 10 minutes to avoid memory errors in the downstream separation stage.
Finally, we drop episodes that yield fewer than four valid dialogue clips, as such episodes are in practice mostly single-speaker content. The output of this stage is a set of two-speaker dialogue clips with diarization labels specifying who spoke when; the two speakers, however, remain mixed into a single channel.

\subsection{Speech Separation and Restoration}
The final stage turns each mixed two-speaker clip into one estimated track per speaker. We separate each retained clip using DialogueSidon~\cite{dialoguesidon}, a diffusion-based model that jointly performs speech separation and restoration (denoising, dereverberation etc.) on degraded in-the-wild dialogue mixtures. The separator maps each monaural two-speaker clip to a pair of speaker-wise waveforms. We store the reconstructed tracks as stereo audio, with one speaker assigned to the left channel and the other to the right channel. This channel-separated representation lets a multi-stream SDLM model the conversational dynamics---overlaps, backchannels, and rapid turn-taking---that full-duplex dialogue requires.

\begin{table}[t]
\centering
\caption{Corpus statistics of DuplexChat. \# stands for ``Number of''}
\label{tab:index-stats}
\footnotesize
\begin{tabular}{lrr}
\toprule
 & Japanese & English \\
\midrule
\# Unique RSS feeds & 8{,}971     & 13{,}041     \\
\# Episodes      & 401{,}494   & 664{,}209    \\
\# Dialogues     & 7{,}329{,}011 & 15{,}304{,}412 \\
Total duration [h]         & 132{,}723   & 282{,}634    \\
Mean dialogue duration [s] & 65.2 & 66.5 \\
\bottomrule
\end{tabular}
\vspace{-4mm}
\end{table}

\begin{table}[t]
\centering
\caption{Audio and separation quality evaluation result. Best value shown in bold. Evaluation was conducted on random 600 samples from each corpus.}
\label{tab:quality}
\footnotesize
\setlength{\tabcolsep}{4pt}
\begin{tabular}{lrrrrr}
\toprule
Corpus & DNSMOS$\uparrow$ & SQ-STOI$\uparrow$ & SQ-PESQ$\uparrow$ & ITC$\uparrow$ & ITD$\uparrow$\\
\midrule
DuplexChat-En & 2.92  & 0.96 & 3.23 & \textbf{0.383} & \textbf{0.880} \\
DuplexChat-Ja & \textbf{3.11}  & \textbf{0.98} & \textbf{3.30} & 0.320 & 0.867 \\
Fisher        & 2.72  & 0.91 & 2.33 & \textbf{0.383} & 0.840\\
\bottomrule
\end{tabular}
\vspace{-4mm}
\end{table}

\begin{table}[t]
\centering
\caption{Turn-taking statistics of DuplexChat}
\label{tab:turntaking}
\begin{tabular}{lrr}
\toprule
Metric & Japanese & English \\
\midrule
Turn exchanges (min$^{-1}$)      & 10.5 & 6.2  \\
Mean turn duration (s)           & 4.0  & 7.9  \\
Backchannels (min$^{-1}$)        & 5.6  & 3.1  \\
Simultaneous speech (\%)         & 21   & 10   \\
Overlapping transitions (\%)     & 50   & 48   \\
\bottomrule
\end{tabular}
\vspace{-4mm}
\end{table}
\section{DuplexChat Statistics and Analysis}
Using DuplexChat-Pipe, we constructed DuplexChat, a corpus of spoken dialogues covering 282{,}634 hours of English and 132{,}723 hours of Japanese two-speaker podcast dialogue. These spans correspond to 14\,TB and 6.7\,TB, respectively. 
For Japanese, we processed all feeds available in the PodcastIndex as of April 2026. Therefore, the reported scale is close to the ceiling achievable from this source. For English, we processed only about $2\%$ of the available RSS feeds due to a computational resource constraint. Therefore, we expect much larger size when the crawl is performed on all available feeds.

Table~\ref{tab:index-stats} shows the statistics of the corpus. DuplexChat covers around 13k unique RSS feed URLs for English and 9k for Japanese, showing the diversity of the corpus.

\subsection{Audio quality evaluation}
The corpus needs to contain acoustically clean speaker-separated dialogues in order to train SDLMs.
To verify this, we compare DuplexChat against Fisher corpus which is widely used in previous SDLM studies~\cite{defossez2024moshispeechtextfoundationmodel,nguyen-etal-2023-generative,roy2026personaplexvoicerolecontrol}.

We report the following metrics:
\begin{itemize}
    \item \textbf{DNSMOS}~\cite{dnsmos}: A machine learning based model which predicts human mean opinion score on acoustic cleanliness.
    \item \textbf{SQ-STOI} and \textbf{SQ-PESQ}: STOI~\cite{5495701} and PESQ~\cite{pesq} values predicted by SQUIM~\cite{10096680} model.
    \item \textbf{ITC}~\cite{shi2025unsupervisedsinglechannelspeechseparation}: Intra track consistency. Cosine similarity of frame-wise speaker embedding extracted from separated output. Higher value indicates consistent speaker identity in the separated track.\footnote{\label{foot:emb}We extracted speaker embedding using  \url{https://hf.co/speechbrain/spkrec-ecapa-voxceleb}}.
    \item \textbf{ITD}~\cite{shi2025unsupervisedsinglechannelspeechseparation}: Inter track distinctiveness. Defined as one minus cosine similarity of speaker embeddings extracted from separated tracks. lower value indicates separation failure.\footref{foot:emb}
\end{itemize}

Table~\ref{tab:quality} presents the results.
From the results, we can see that DuplexChat has better acoustic cleanliness compared to the Fisher corpus. For speaker separation (ITC and ITD), the results show that DuplexChat-En achieves separation quality comparable to Fisher, whereas DuplexChat-Ja is lower.
This can be due to the limited Japanese training data of DialogueSidon which performs separation on monaural dialogue.

\subsection{Turn-Taking Dynamics}
An ideal corpus for SDLM training should reflect human turn-taking dynamics. To analyze the turn-taking dynamics of DuplexChat, we measure turn-taking statistics on a random sample of roughly 700 (English) and 900 (Japanese) dialogues. We follow the turn-taking analysis of \textit{Full-Duplex-Bench} and adapt its thresholds to our separated channels. Specifically, we run a voice-activity detection~\cite{silerovad} on each channel and merge segments separated by less than $0.5$\,s into \emph{talk spurts}, taking word counts from ASR models\footnote{For English we used \url{https://hf.co/nvidia/parakeet-tdt-0.6b-v2}. For Japanese we used \url{https://huggingface.co/nvidia/parakeet-tdt_ctc-0.6b-ja}}. A talk spurt of at least $1$\,s, or with more than three words, is a turn; a sub-second, few-word spurt contained within the other speaker's turn is a backchannel.

We report the following metrics:
\begin{itemize}
  \item \textbf{Turn exchanges} (min$^{-1}$): Number of turn exchanges from one speaker to the other, per minute.
  \item \textbf{Mean turn duration} (s): average length of a turn.
  \item \textbf{Backchannels} (min$^{-1}$): backchannels per minute.
  \item \textbf{Simultaneous speech} (\%): fraction of time both channels are active at once.
  \item \textbf{Overlapping transitions} (\%): percentage of turn taking with overlap.
\end{itemize}

Table~\ref{tab:turntaking} reports the results for this analysis.
We can see that while overlapping transitions are similar between the two languages, in other metrics the values differs substantially. Specifically, Japanese dialogues show more frequent turn exchanges and backchannels, and contain more simultaneous speech.
This result is consistent with the previous analyses of turn-taking dynamics~\cite{liberman2007towards}, confirming that DuplexChat captures the turn-taking dynamics present in spoken dialogues.


\section{Conclusion}
We presented DuplexChat-Pipe, an open-source and re-runnable
pipeline for constructing speaker-separated full-duplex dialogue
speech from public podcast feeds, and DuplexChat, the corpus
constructed using it. DuplexChat-Pipe combines feed-level filtering,
audio retrieval and cleaning, diarization-based dialogue segmentation,
and speech separation/restoration to estimate one track per speaker.
Using this pipeline, we constructed 282,634 hours of English and
132,723 hours of Japanese two-speaker podcast dialogue. Our analyses
show that DuplexChat achieves competitive acoustic quality
and speaker-consistency scores relative to Fisher, and that it exhibits
key conversational patterns, including backchannels, overlapping
speech, and frequent turn exchanges. Future work will evaluate how
DuplexChat affects downstream spoken dialogue language model training
and how corpus scale influences full-duplex dialogue modeling.

\section*{Acknowledgments}
This work was supported by JST Moonshot JPMJMS2011, JST BOOST JPMJBY24C9, JSPS KAKENHI, Grant Number 25KJ0806, and the AIST policy-based budget project ``R\&D on Generative AI Foundation Models for the Physical Domain.''

\section*{Use of Generative AI disclosure}
Claude Opus 4.8 was used for implementation and to improve the textual content of this paper.
\section*{Ethics statement}
Release of the crawled audio can result in copyright issues. Therefore, the released dataset only includes audio URLs and dialogue segment durations to avoid such issues. We also release the code to reconstruct the dataset using the released metadata. When using this dataset we ask the users to respect the laws and rights of original copyright holders. Furthermore, we implement opt-out policy that allows individuals or rights holders to request removal of their data from the dataset.
\bibliographystyle{IEEEbib}
\bibliography{custom}

\end{document}